%% file: 0_main.tex
\documentclass[11pt,a4paper]{article}
\usepackage[hyperref]{acl2021}
\usepackage{times}
\usepackage{latexsym}

\usepackage{microtype}

\aclfinalcopy 
\def\aclpaperid{3272} 

\usepackage{amsmath}
\usepackage{amsfonts}
\usepackage{xcolor}
\usepackage{xspace}
\usepackage{multirow}
\usepackage{paralist}
\usepackage{graphicx}
\usepackage{subcaption}
\usepackage{float}
\usepackage{subcaption}
\usepackage{tabularx}
\usepackage{booktabs}

\newcommand\sect[1]{\S\ref{#1}}
\newcommand\septoken{{\small\textsc{[SEP]}}\xspace}
\newcommand\clstoken{{\small\textsc{[CLS]}}\xspace}
\DeclareMathOperator{\softmax}{Softmax}

\DeclareMathOperator{\rank}{rank}
\DeclareMathOperator{\nullspace}{LN}
\DeclareMathOperator{\myspan}{span}
\DeclareMathOperator{\proj}{P}

\title{Effective Attention Sheds Light On Interpretability}

\author{Kaiser Sun\textsuperscript{*} \quad Ana Marasovi\'{c}\textsuperscript{$\dagger$*} \\ 
\textsuperscript{$*$}Paul G.\ Allen School of Computer Science \& Engineering, University of Washington\\
\textsuperscript{$\dagger$}Allen Institute for AI\\
Seattle, WA, USA \\
\texttt{huikas@cs.washington.edu}, \texttt{anam@allenai.org}}
  
\date{}

\begin{document}
\maketitle
\begin{abstract}
An attention matrix of a transformer self-attention sublayer can provably be decomposed into two components and only one of them (\emph{effective attention}) contributes to the model output. This leads us to ask whether visualizing effective attention gives different conclusions than interpretation of standard attention. Using a subset of the GLUE tasks and BERT, we carry out an analysis to compare the two attention matrices, and show that their interpretations differ. Effective attention is less associated with the features related to the language modeling pretraining such as the separator token, and it has more potential to illustrate linguistic features captured by the model for solving the end-task. Given the found differences, we recommend using effective attention for studying a transformer's behavior since it is more pertinent to the 
model output by design.
\end{abstract}

\input{1_introduction}

\input{2_background}

\input{3_comparison}

\input{4_conclusions}


\section*{Acknowledgments}
The authors thank Noah A.\ Smith, members of Noah's ARK, as well as anonymous reviewers for their helpful feedback, and Olga Kovaleva for sharing the code and model weights for classification of attention patterns.

\bibliography{anthology, acl2021}
\bibliographystyle{acl_natbib}
\clearpage
\appendix
\input{appendix}

\end{document}

%% file: 1_introduction.tex
\section{Introduction}

Attention mechanism \cite{Bahdanau2015NeuralMT} is an essential component of many NLP models, including those  that are built on the ubiquitous transformer architecture \cite{vaswani2017attention}.  
As a result, visualizing attention weights is a widely used technique to interpret models' behavior \cite{belinkov-glass-2019-analysis}. Despite that, the validity of this analysis method is a subject undergoing intense discussion and study in NLP \cite[\emph{i.a.}]{jain-wallace-2019-attention, wiegreffe-pinter-2019-attention, serrano-smith-2019-attention, moradi-etal-2019-interrogating, mohankumar-etal-2020-towards, tutek-snajder-2020-staying}. 

Related to this discussion, \citet{Brunner2020OnII} show that, under mild conditions, the attention matrix of a transformer self-attention sublayer can be written as a sum of two components. One of them is irrelevant for the model output because its product with the value matrix is zero. They term the other component as  \emph{effective attention} (formally defined in \sect{sec:background}). We study whether effective attention gives interpretations that differ from conclusions we get by analyzing standard attention. If this is the case, interpretation of effective attention is better suited for studying transformers' internals because it is more pertinent to the model output by design. 


\citet{Brunner2020OnII} briefly discuss this by comparing  standard and effective attention matrices from a single BERT  head \cite{devlin-etal-2019-bert} for one example. They observe that: (i) standard attention is largely concentrated on the delimiter tokens (\septoken, \clstoken) or on near-diagonal elements; (ii) effective attention is more dispersed; (iii) effective attention disregards the delimiters. They stress that we should not extrapolate too much from these observations since they are based on a single example, and that further research is needed on this topic.  

In this work, we aim to reliably answer whether effective attention disregards the \septoken and \clstoken tokens, and if so, are effective attention weights dispersed to linguistic features? To address these questions, we embrace the methodology for a quantitative analysis of the attention patterns produced by individual transformer heads proposed by \citet{kovaleva-etal-2019-revealing}. We carry out their experiments on a subset of the GLUE tasks with BERT's standard and effective attention. We show that effective attention ``ignores'' \septoken and punctuation symbols (\sect{sec:patterns_classification}, \sect{sec:attention_features}), but not \clstoken  (\sect{sec:attention_features}), and that it highlights  end-task features instead (\sect{sec:patterns_classification}, \sect{sec:attention_features}, \sect{sec:finetuning}).\footnote{Our code is available at \url{https://github.com/KaiserWhoLearns/Effective-Attention-Interpretability}}



%% file: 2_background.tex
\section{Background: Effective Attention}
\label{sec:background}

Each transformer layer consists of multi-head self-attention and feedforward sublayers \cite[see Appendix \ref{sec:appendix_background}]{vaswani2017attention}. %
\citet{Brunner2020OnII} show that the \textbf{standard attention} matrix $A$ can be decomposed into two components, if a mild condition is satisfied. Specifically, if the left nullspace of the value matrix $V$: 
\begin{equation*}
    \nullspace(V):=\{x^\top \in \mathbb{R}^{1\times d_s} | x^\top V=0\},
\end{equation*}
is not trivial (contains vectors other than $\vec{0}$). This is satisfied when the maximum input sequence length is larger than the value matrix dimension (see Appendix \ref{sec:appendix_background}). The two components are: the component in the left nullspace of $V$ ($A^{\parallel}$) and the component orthogonal to the nullspace ($A^{\perp}$). %
Notably, $A^{\parallel}$ does not contribute to the output of the self-attention sublayer:
\begin{equation}
    AV = (A^{\parallel} + A^{\perp})V = \vec{0} + A^{\perp}V=A^{\perp}V. \label{eq:contribution}
\end{equation}
The \textbf{effective attention} matrix is defined as $A^{\perp}$. If visualizations of standard and effective attention differ, interpretation of effective attention is an accurate interpretation because effective attention is what contributes to the model output (per Eq.\ \ref{eq:contribution}). 

We explain how to compute $A^{\perp}$ since that was not described in \citet{Brunner2020OnII}. We first compute the singular value decomposition (SVD) of the value matrix $V=U\Sigma W^{T}$. The rows of U that correspond to singular values equal to zero span $\nullspace(V)$: 
\begin{equation*}
  \nullspace(V)= \myspan\{u_1,\hdots,u_k\},
\end{equation*}
where $k$ is the number of singular values that equal zero. We project each row $a_i$ of the attention matrix $A \in \mathbb{R}^{d_s \times d_s}$ to $\nullspace(V)$ to construct a projection of the \emph{matrix} $A$ to $\nullspace(V)$:
\begin{align*}
    \proj_{\nullspace(V)}(a_i) &=\sum_{j=1}^{k}\langle a_i,u_j\rangle u_j, \forall i \in \{1,\hdots, d_s\},\\
    \proj_{\nullspace(V)}(A) &= [\proj_{\nullspace(V)}(a_1), \hdots, \proj_{\nullspace(V)}(a_{d_s})]^\top,
\end{align*}

where $\langle \cdot,\cdot\rangle$ denotes the dot product. Finally, effective attention equals to:

\begin{equation*}
   A^{\perp} := A - \proj_{\nullspace(V)}(A). 
\end{equation*}

Effective attention is not guaranteed to be a probability distribution as some of its weights might be negative and larger than 1.

We observe that effective attention is slower to compute due to the SVD decomposition of $V$ for each out of 144 BERT-base heads, and additional matrix multiplications (Table \ref{tab:evaluation-time}; \sect{sec:appendix_results}). If speed is bottleneck, we recommend doing quantitative analyses with effective attention on a subset of the dev set. For qualitative analyses, common practice is already to select a subset for a manual analysis.

%% file: 3_comparison.tex
\section{What Does Effective Attention Reveal?}
\label{sec:experiments}

We compare visualizations of standard and effective attention following the methodology for analysis of the attention patterns \cite{kovaleva-etal-2019-revealing}. We carry out our analyses using five English-language datasets in the GLUE benchmark \citep{wang2019glue}: RTE \cite{dagan2005pascal,haim2006second,giampiccolo2007third,bentivogli2009fifth}, MRPC \cite{dolan2005automatically}, QNLI \cite{rajpurkar-etal-2016-squad,wang2019glue}, SST-2 \cite{socher-etal-2013-recursive}, and STS-B \cite{cer-etal-2017-semeval}.\footnote{We omit larger datasets (QQP, MNLI), due to the limit of our computation budget (a single Nvdia GTX1070 with 8GB  memory), and CoLA/WNLI following \citet{kovaleva-etal-2019-revealing}.} See Table \ref{tab:datasets} for their specifications. For each dataset, we train BERT-base with standard attention, a batch size of 8, maximum sequence length of 128, and 3 training epochs.\footnote{All other hyperparameters are set to default values in the transformers library \cite{wolf-etal-2020-transformers}.} For analyzing effective attention, we replace standard with effective attention at the test time.

\input{tables/data_stats}

\subsection{Classification of Attention Patterns}
\label{sec:patterns_classification}

In this section, we start studying whether effective attention disregards the delimiter tokens. 

The visualizations of attention matrices exhibit patterns \cite{clark-etal-2019-bert, vig-belinkov-2019-analyzing}. \citet{kovaleva-etal-2019-revealing} identified five frequently occurring pattern categories: 
\begin{compactitem}
\item vertical (associated with the \emph{delimiters tokens})
\item diagonal (either syntactic features between neighbouring words in the English language or the previous/following token attention coming from the language modeling  pretraining)
\item vertical + diagonal 
\item block (intra-sentence attention for the tasks with two distinct sequences; potentially encodes semantic and syntactic information)
\item heterogeneous (as ``block'', more likely to capture interpretable linguistic features).
\end{compactitem}

They annotated 400 BERT's attention matrices using these categories, and used them to train a ConvNet for pattern classification of 1K random test set attention matrices. We replicate their results for standard attention (using their code), and classify effective attention matrices for a comparison.\footnote{We thank the authors for sharing their code and model weights for this experiment.}

\paragraph{Results}

Table \ref{tab:attention-patterns} (Fig.\ \ref{fig:classify} in Appendix \ref{sec:appendix_results}) shows a drop in the percentage of the ``vertical'' and ``vertical + diagonal'' patterns when we replace the standard with effective attention. Since the vertical patterns are associated predominantly with attention to the delimiters tokens, this result supports the hypothesis that effective attention disregards the delimiter tokens. Moreover, although the amount of ``heterogeneous'' patterns did not change notably, the amount of ``block'' and ``diagonal'' patterns increased. This suggests that we are better positioned to find end-task linguistic features captured by the model by visualizing effective attention.

As an illustration, Figure \ref{fig:example} presents the attention matrices for one sentence from one attention head. In this example, effective attention highlights all mentions of the noun ``antibiotics'' that the adjective ``new'' modifies and that is also the object of the preposition ``against'', instead of giving prominence to the \septoken  token as standard attention.

\input{tables/attention_patterns}
\input{figs/tags}
\input{figs/example_figs/example}

\subsection{Delimiter Tokens vs.\ Linguistic Features}
\label{sec:attention_features}

We showed that the ``vertical'' pattern, \emph{associated} with the delimiter tokens, is less dominant with effective attention (\sect{sec:patterns_classification}). To verify that both delimiter tokens are indeed less relevant with effective attention, following \citet{kovaleva-etal-2019-revealing}, we report the standard and effective attention weights of specific token types when processing the \clstoken token in the final layer. Namely, the attention weights of linguistic features (nouns, pronouns, verbs), the delimiter tokens (\septoken, \clstoken), and punctuation symbols that are conceptually similar to \septoken.\footnote{If there are multiple tokens of the same type in the input, we use the one with the maximum weight. If a word consists of the multiple subtokens, we use the  weight of the first subtoken.}

\paragraph{Results} Figure \ref{fig:tags} shows that \septoken is among the two most relevant features for all tasks except \textsc{\small QNLI} according to standard attention (upper two rows in each subfigure, colored green). For all but one task (\textsc{\small SST-2}), it loses its dominance with effective attention and its weights are apparently shifted to linguistic features. This is also the case for punctuation symbols. This result shows that the \septoken token and punctuation symbols are not as important for understanding how the model solves the end-task as standard attention suggests.  

We observe that \clstoken is attended similarly with effective and standard attention, contrary to what \citeauthor{Brunner2020OnII} suggested. To rule out this is because we plot the attention assigned to \clstoken when processing \clstoken, we report the attention  assigned to \clstoken when processing other input words (regardless of their type) in Fig.\ \ref{fig:cls_token} in Appendix \ref{sec:appendix_results}. Again, we do not observe differences between standard and effective attention, unlike for  \septoken (Fig.\ \ref{fig:sep_token} in \sect{sec:appendix_results}). These results confirm the hypothesis of \citeauthor{Brunner2020OnII} that effective attention disregards \septoken, but not  \clstoken as they also hypothesized. Notably, \septoken is associated with the LM pretraining and \clstoken only with the task-specific finetuning. 

\subsection{Effects of Task-Specific Finetuning}
\label{sec:finetuning}

\input{figs/cos_similarity}

To provide our final evidence that effective attention captures end-task features, we investigate how attention changes with finetuning layer-wise; again following \citet{kovaleva-etal-2019-revealing}. 
They calculate the cosine similarity between pretrained and finetuned flattened attention matrices. The layers that change the most, encode most task-specific features. 
To reiterate, effective attention is the part of standard attention that  contributes to the model output (Eq.\ \ref{eq:contribution}; \sect{sec:background}), and we showed that it is less associated with the pretraining feature \septoken and more with linguistic features (\sect{sec:patterns_classification}, \sect{sec:attention_features}). Thus, changes of standard attention from task-specific finetuning should be the product of changes of effective attention, and the outcome of this analysis should be the same, regardless of the attention ``type''.

\paragraph{Results}
As expected, we come to the same conclusion with effective attention as \citeauthor{kovaleva-etal-2019-revealing} did with the standard: the last two layers change the most with finetuning (Fig.\ \ref{fig:cos_sim}). This soundness check suggests once again that effective attention is the component of standard attention that manifests end-task features. 


%% file: tables/data_stats.tex
\begin{table}[t]
\resizebox{\columnwidth}{!}{
\begin{tabular}{llll}
\toprule
\textbf{Dataset} & \textbf{Task}                             & \textbf{$|$Train$|$} & \textbf{$|$Test$|$} \\
\midrule
RTE     & NLI       & 2.5K    & 3K     \\
MRPC    & paraphrase identification        & 3.7K    & 1.7K   \\
QNLI    & QA as NLI & 105K    & 5.4K   \\
SST-2   & binary sentiment classification  & 67K     & 1.8K   \\
STS-B   & sentence similarity              & 7K      & 1.4K  \\
\bottomrule
\end{tabular}
}
\caption{Specifications of the datasets.}
\label{tab:datasets}
\end{table}

%% file: tables/attention_patterns.tex
\begin{table}[t]
\centering
\resizebox{\columnwidth}{!}{
\begin{tabular}{llrrrrr}
\toprule
\textbf{Task} & \textbf{Attention} & \textbf{B} & \textbf{D} & \textbf{V+D} & \textbf{H} & \textbf{V} \\ 
\midrule
& Standard                                                            & 4.50                                                        & 7.40                                                           & 15.20                                                             & 45.10                                                               & 27.90                                                          \\ 
\multirow{-2}{*}{RTE}                                            & Effective                                   & 32.60                               & 12.80                                  & 2.80                                      & 40.30                                       & 11.50                                  \\ \midrule
                                       & Standard                                                            & 3.40                                                        & 10.20                                                          & 14.90                                                             & 39.80                                                               & 31.80                                                          \\ 
\multirow{-2}{*}{MRPC}                   & Effective                                                           & 25.50                                                       & 17.40                                                          & 3.60                                                              & 40.40                                                               & 13.00                                                          \\ \midrule
                                                                 & Standard                                                            & 4.70                                                        & 7.40                                                           & 15.20                                                             & 45.10                                                               & 27.90                                                          \\ 
\multirow{-2}{*}{QNLI}                                           & Effective                                   & 29.30                               & 15.80                                  & 3.40                                      & 46.40                                       & 5.10                                   \\ \midrule
                                         & Standard                                                            & 38.50                                                       & 6.10                                                           & 0.00                                                              & 37.80                                                               & 17.60                                                          \\ 

\multirow{-2}{*}{SST-2}                                          & Effective                                   & 33.80                               & 11.50                                  & 0.80                                      & 39.40                                       & 14.60                                  \\ \midrule
                                        & Standard                                                            & 4.00                                                        & 8.20                                                           & 1.80                                                              & 50.40                                                               & 35.50                                                          \\ 
\multirow{-2}{*}{STS-B}                  & Effective                                                           & 36.00                                                       & 10.30                                                          & 0.60                                                              & 39.40                                                               & 13.60                                                          \\ 
\bottomrule
\end{tabular}}
\caption{Estimated percentage of the attention patterns (\sect{sec:patterns_classification}): block (B), diagonal (D), vertical + diagonal (V + D), heterogeneous (H), vertical (V). Effective attention exhibits different patterns than standard attention, i.e., less vertical patterns (associated with delimiter tokens) and more block patterns (associated with task features).}

\label{tab:attention-patterns}
\end{table}

%% file: figs/tags.tex

\begin{figure}[!h]
    \centering
    \subfloat[\textsc{RTE}]{\includegraphics[width=0.47\textwidth]{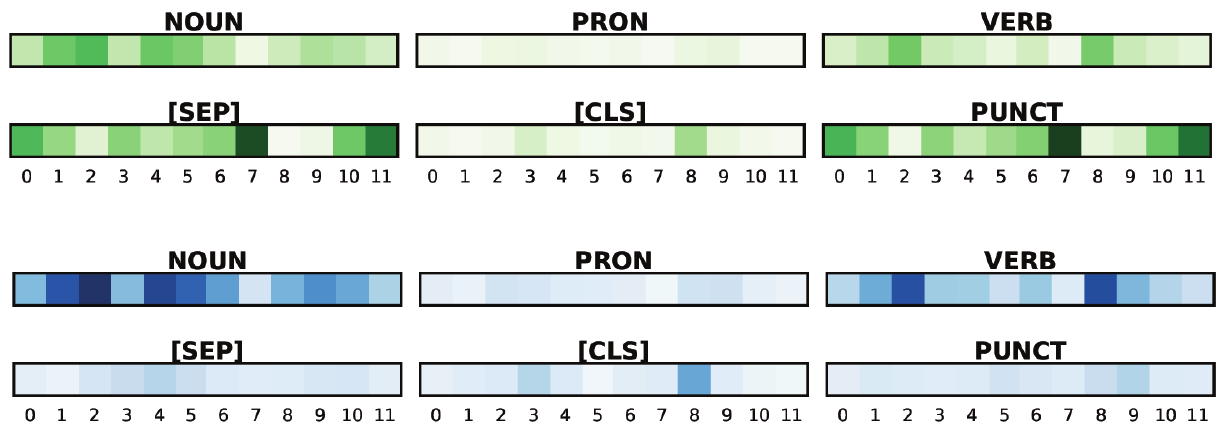} }
    \qquad
    \subfloat[\textsc{MRPC}]{\includegraphics[width=0.47\textwidth]{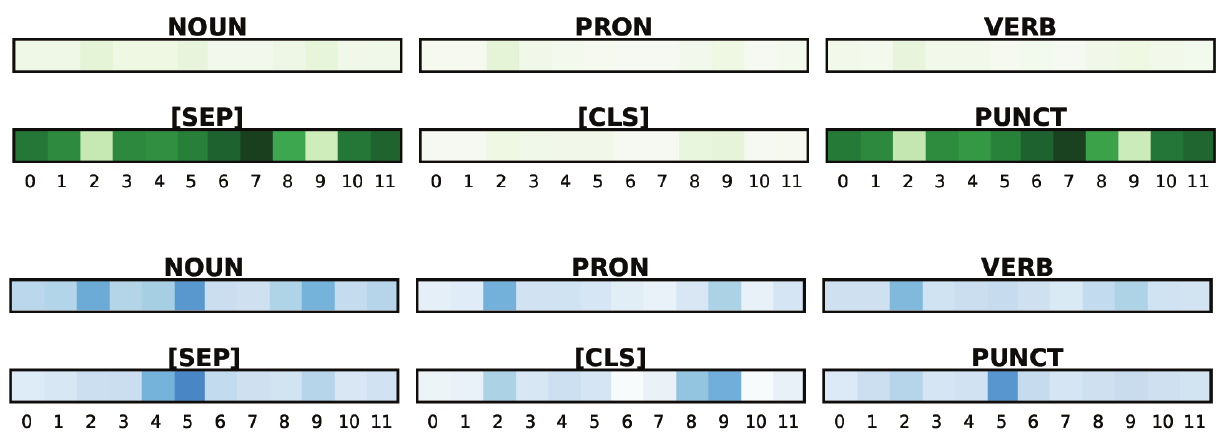} }
    \qquad
    \centering
    \subfloat[\textsc{QNLI}]{\includegraphics[width=0.47\textwidth]{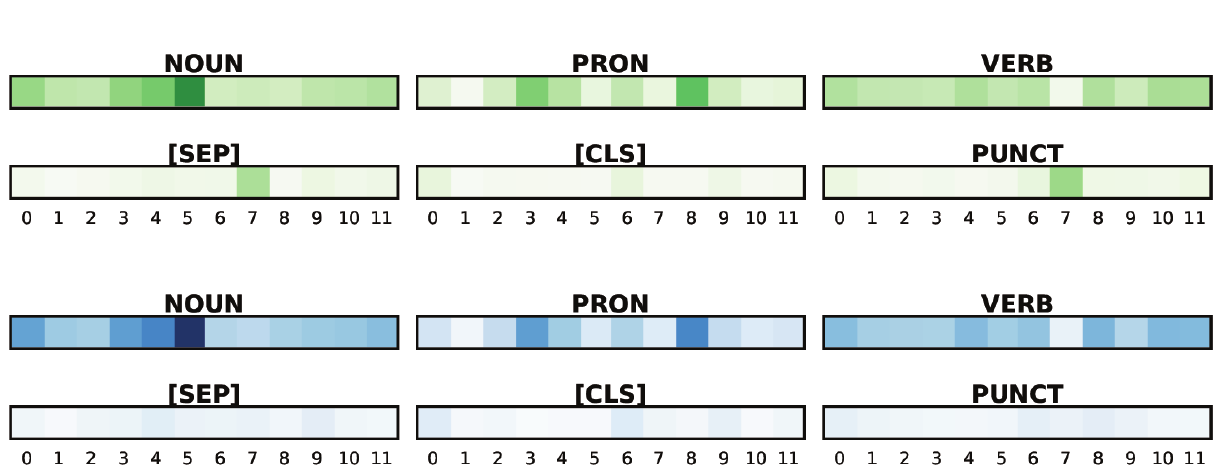} }
    \qquad
    \subfloat[\textsc{SST-2}]{\includegraphics[width=0.47\textwidth]{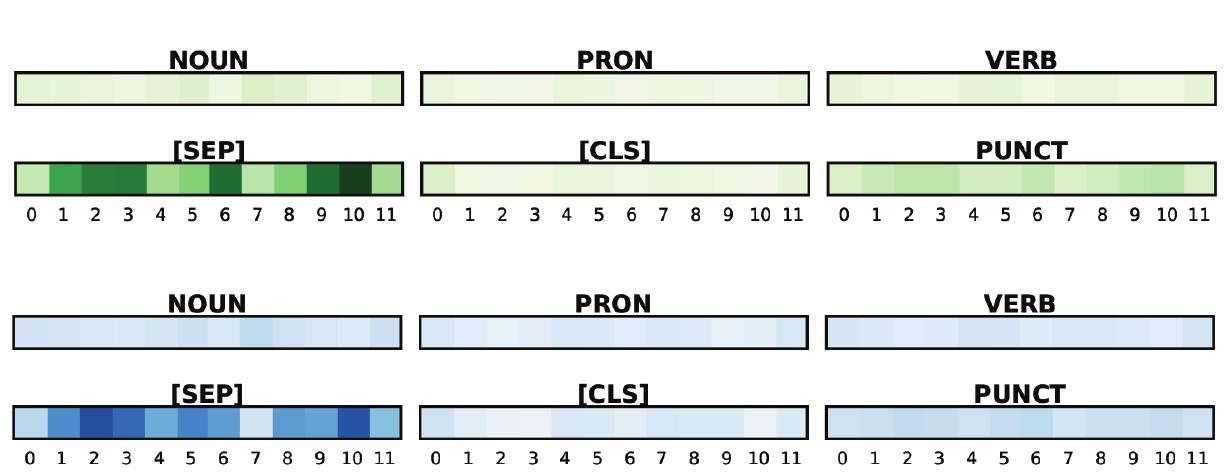} }
    \qquad
    \centering
    \subfloat[\textsc{STS-B}]{\includegraphics[width=0.47\textwidth]{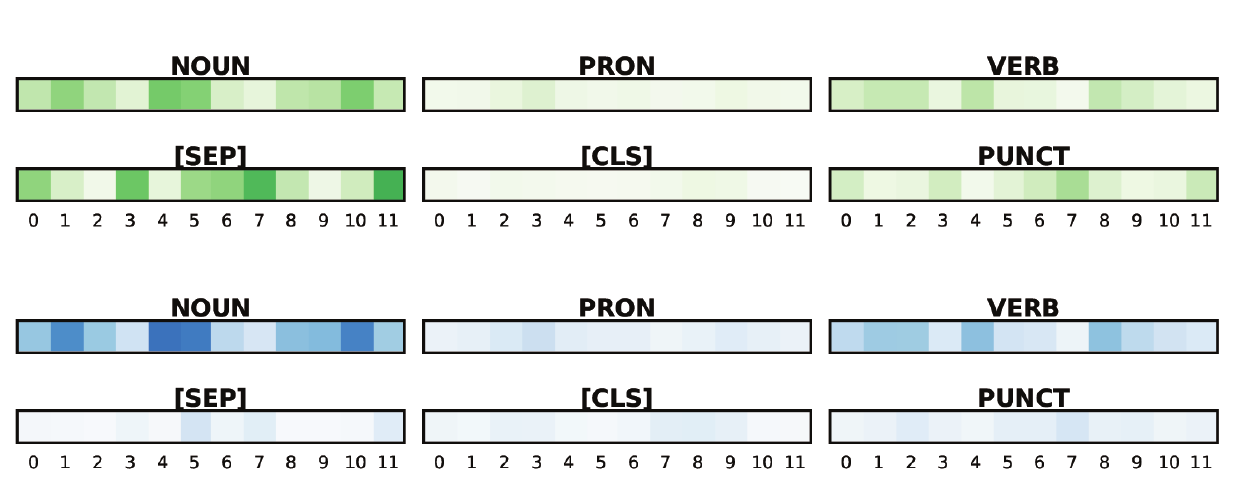} }
    \qquad
    \centering
    \subfloat[\textsc{Average over Tasks}]{\includegraphics[width=0.47\textwidth]{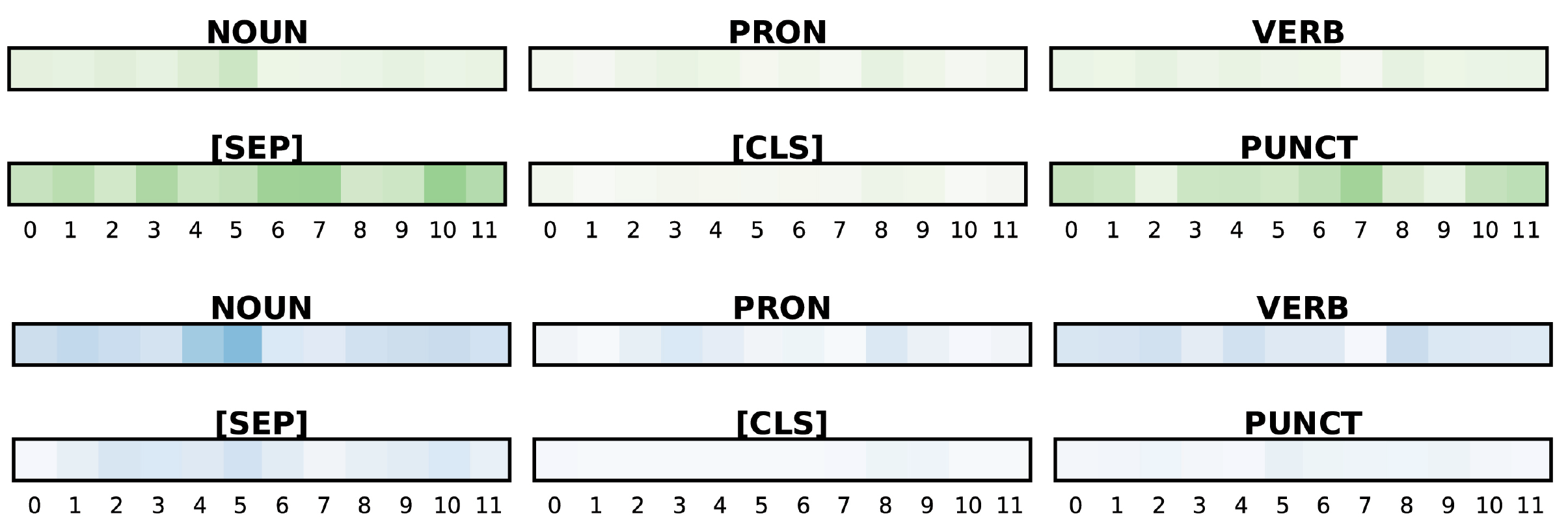} }
    \caption{Effective attention ``pays less attention'' to \septoken and punctuation. Per-task and per-head (0--11) attention when processing \clstoken in the final layer, averaged over test set. The darker colors correspond to larger attention values. The green plots (two upper rows in subfigures) illustrate standard, and blue plots (two lower rows in subfigures) effective attention. }
    \label{fig:tags}
\end{figure}

%% file: figs/example_figs/example.tex
\begin{figure*}[t]
    \centering
    \subfloat[Standard attention]{\includegraphics[width=0.7\textwidth]{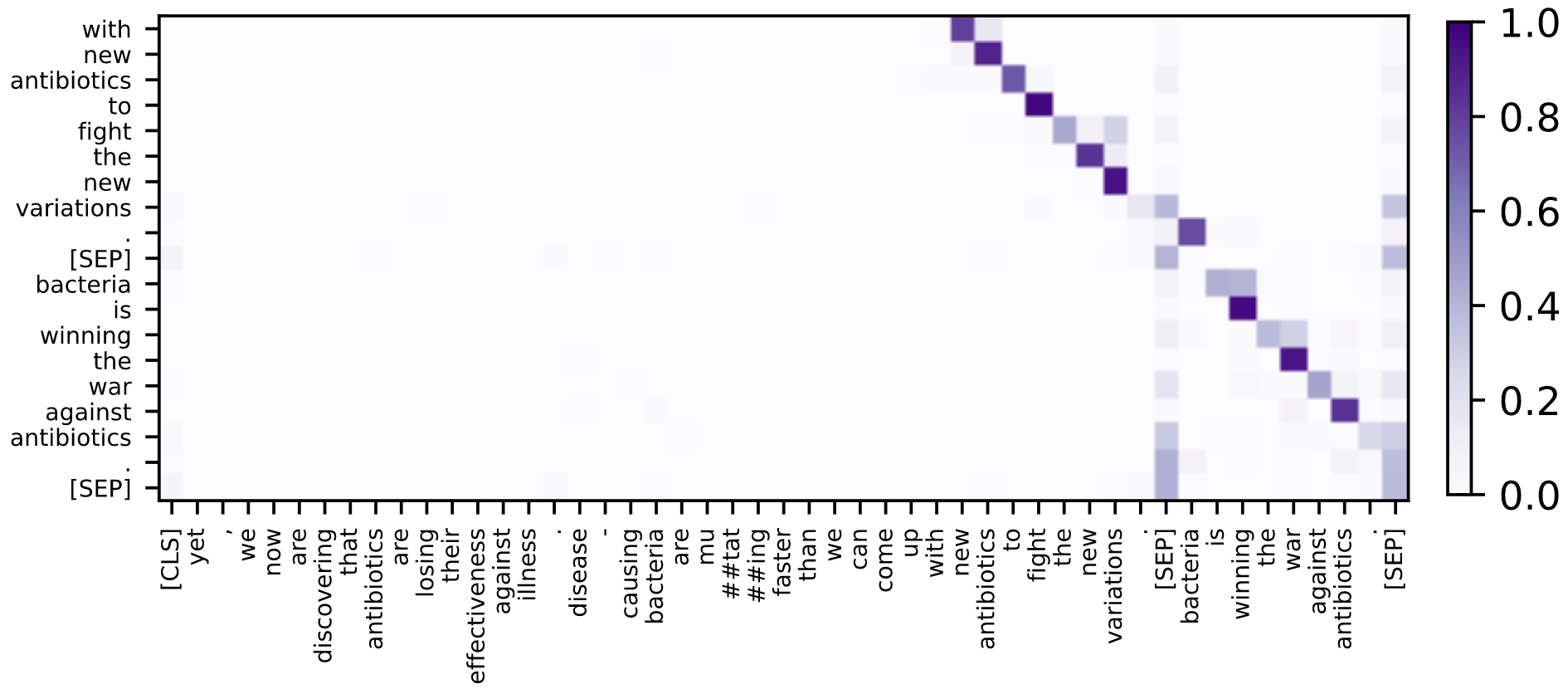} }
    \quad
    \centering
    \subfloat[Effective attention]{\includegraphics[width=0.7\textwidth]{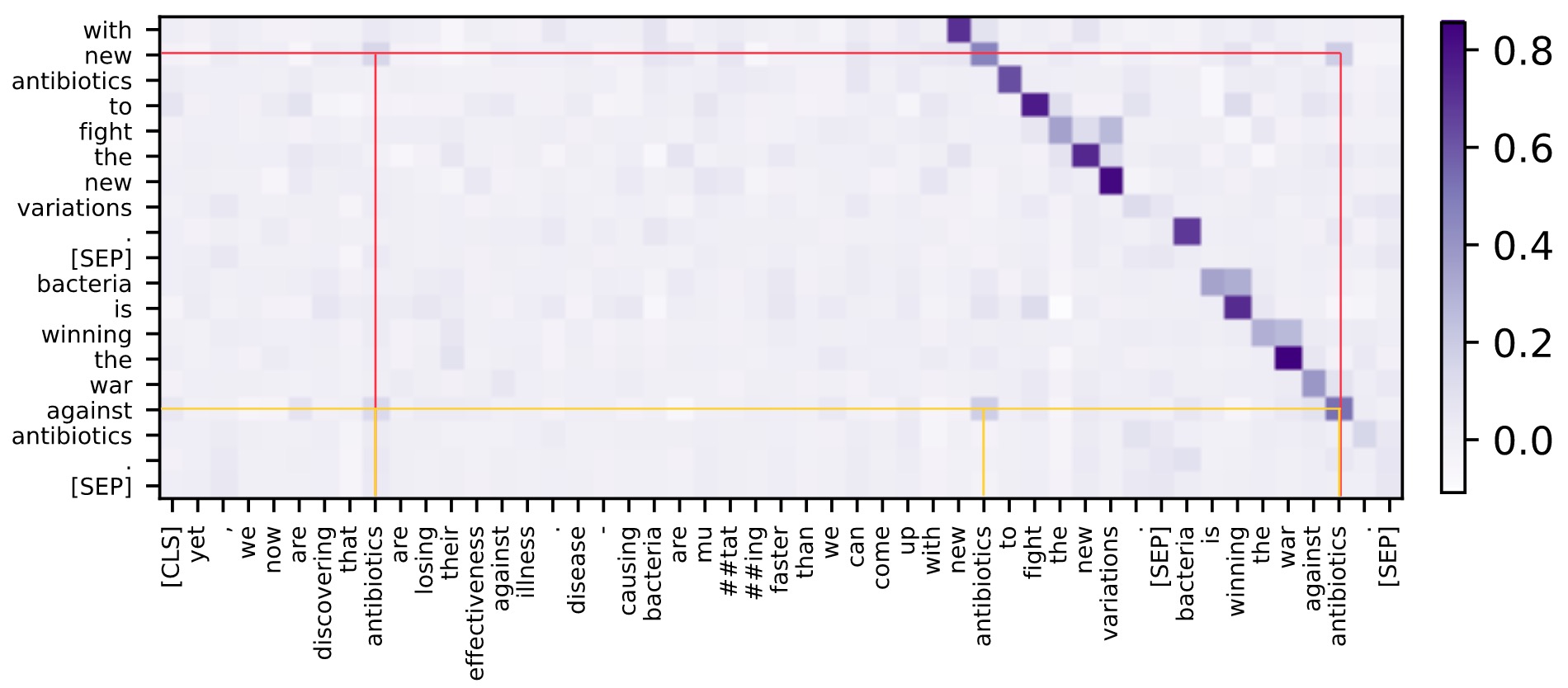} }
    \caption{Visualizations of standard and effective attention from one head for one example from the \textsc{\small RTE} dataset (recognizing textual entailment). 
    Only the last few rows are visible; see the full version in Fig.\ \ref{fig:example_full} (Appendix \sect{sec:appendix_results}).}
    \label{fig:example}
\end{figure*}

%% file: figs/cos_similarity.tex
\begin{figure*}[t]
\centering
\includegraphics[width=\the\textwidth]{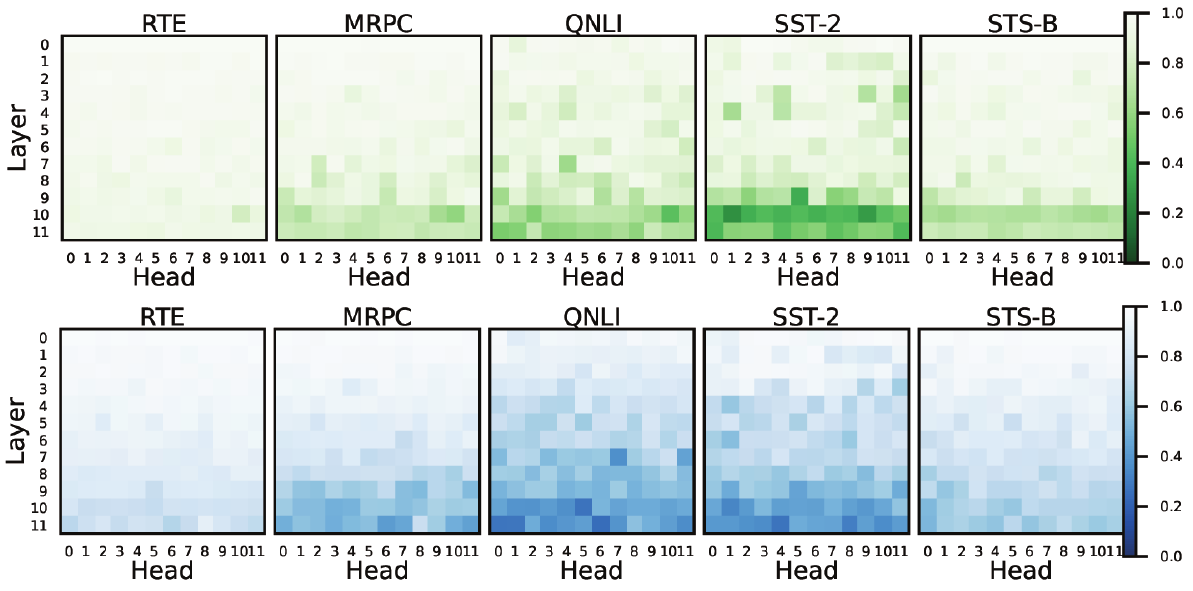}
\caption{Per-task cosine similarity between the pretrained and finetuned attention weights for selected GLUE tasks, calculated across layers and heads. The darker colors corresponding to larger absolute attention weights. The top (green) figure is computed with the standard attention, and the bottom (blue) figure with the effective attention.}
\label{fig:cos_sim}
\end{figure*}

%% file: 4_conclusions.tex
\section{Conclusions}
We study whether effective attention, the part of the transformer attention matrix that does not get canceled out with the value matrix, gives different interpretations than standard attention. We present a comparison of the two attentions and show that they  differ in weights assigned to delimiter tokens such as \septoken and punctuation marks, but not \clstoken as it was previously thought. Instead, effective attention gives more weight to linguistic features. Given the differences, and that effective attention is more pertinent to the model output by design, we urge to use it for studying transformers' internals.

As an alternative to analyzing attention \emph{weights}, \citet{kobayashi-etal-2020-attention} propose anayzing the norm of vectors produced by multiplying the outputs of the value matrix with the attention weights. Following the experimental setting of \citet{clark-etal-2019-bert}, i.e., by analyzing  992 sequences extracted from Wikipedia, their norm-based analysis also shows that the contributions of \septoken and punctuations are actually small. However, unlike us, they report the same observation for \clstoken. Future work might consider a more formal study between the norm-based analysis and effective attention, especially since the norm-based analysis could circumvent the problem of costly SVD.

%% file: appendix.tex
\section{Background: On The Rank Of The Value Matrix}
\label{sec:appendix_background}
The output $Z$ of an individual self-attention head is given by:
\allowdisplaybreaks
\begin{align*}
Q &= Z_{l-1}W^{Q} \in \mathbb{R}^{d_s \times d_q} \\
K &= Z_{l-1}W^{K} \in \mathbb{R}^{d_s \times d_k} \\
V &= Z_{l-1}W^{V} \in \mathbb{R}^{d_s \times d_v} \\
A &= \softmax\Big(\frac{QK^T}{\sqrt{d_k}}\Big) \in \mathbb{R}^{d_s \times d_s}\\
Z &= AV \in \mathbb{R}^{d_s \times d_v},
\end{align*}

where $d_s$ is the maximum length of the input sequence (in number of subtokens), $Z_{l-1}$ 
is the output of the previous transformer layer, 
$W^{Q}, W^{K}, W^{V}$ are the query, key, and value \emph{weight} matrices, respectively. For BERT-base, $d_q=d_k=d_v=64$, $n_{\text{heads}}=12$, $d_s=512$, and $d_v \cdot n_{\text{heads}}=768$.

\citet{Brunner2020OnII} show that the upper bound of the rank of the value matrix $V$ is given by:
\begin{align*}
 \rank(V)&=\rank(Z_{l-1}W^{V}) \\
         &\leq \min\{d_s,d_v,d_s,d_v \cdot n_{\text{heads}}\} \\
         &\leq \min\{d_s, d_v\}.   
\end{align*}
As a result, the left nullspace of $V$, defined as:
\begin{equation*}
\nullspace(V):=\{x^\top \in \mathbb{R}^{1\times d_s} | x^\top V=0\},    
\end{equation*}
is non-trivial ($\nullspace(V) \neq \{\vec{0}\}$) when the maximum input length, $d_s$, is larger than the dimension of the value matrix $d_v$, i.e., $d_s > d_v$. In this case, we can construct infinitely many matrices $A+\tilde{A}$,
\begin{equation*}
    \tilde{A}=[x_1,\hdots,x_{d_s}]^\top, x_i \in \nullspace(V),
\end{equation*}
which contribute exactly the same to the output as the attention matrix A:
\begin{equation*}
    (A+\tilde{A})V=AV+\tilde{A}V=AV + \vec{0} = AV.
\end{equation*}
This also holds when the weights of $A+\tilde{A}$ are constrained to the probability simplex, and such constrained matrices $A+\tilde{A}$ exist.

\section{Additional Results}
\label{sec:appendix_results}

We provide the following additional results that complement the discussions in Section \ref{sec:experiments}:
\begin{compactitem}
\item A comparison of the evaluation time with standard vs.\ effective attention. 
\item In Figure \ref{fig:classify}, visualization of results presented in Table \ref{tab:attention-patterns}.
\item Attention to the \clstoken token in Figure \ref{fig:cls_token}.
\item Attention to the \septoken token in Figure \ref{fig:sep_token}.
\item Complete Figure \ref{fig:example}.
\end{compactitem}
\input{tables/eval_time}
\input{figs/attention_patterns_classification}

\input{figs/cls_token}

\input{figs/sep_token}

\input{figs/example_figs/example_full}

%% file: tables/eval_time.tex
\begin{table}[t]
\centering
\resizebox{\columnwidth}{!}{
\begin{tabular}{lcccccc}
\toprule
& \textbf{RTE} & \textbf{MRPC} & \textbf{QNLI} & \textbf{SST-2} & \textbf{SST-B} \\
\midrule
standard & 0:29 & 0:45 & 10:59 & 1:41 & 2:54 \\
effective & 0:58 & 1:27 & 21:05 & 3:20 & 5:53 \\
\bottomrule
\end{tabular}}
\caption{A comparison of the evaluation clock time (minutes:seconds) of BERT models (trained with the standard attention) evaluated with standard attention and  effective attention separately. 
}
\label{tab:evaluation-time}
\end{table}

%% file: figs/attention_patterns_classification.tex
\begin{figure}[t]
\centering
\begin{subfigure}[t]{0.92\columnwidth}
\centering
\includegraphics[width=\columnwidth]{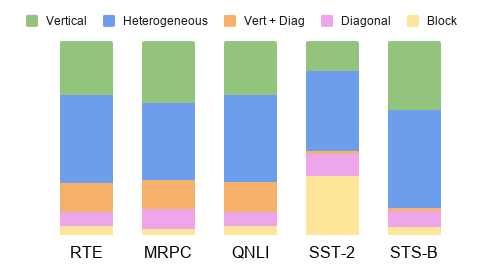}
\caption{Standard attention.}
\label{fig:patterns_standard_attention}
\end{subfigure}\text{ }
\begin{subfigure}[t]{\columnwidth}
\centering
\includegraphics[width=0.92\columnwidth]{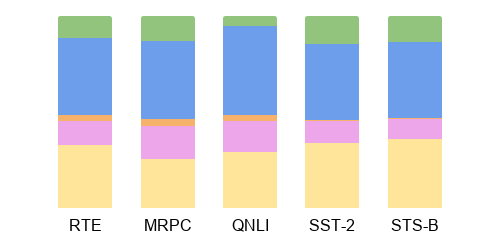} 
\caption{Effective attention.}
\end{subfigure}
\caption{Estimated percentage of the attention patterns (\sect{sec:patterns_classification}) for each task.}
\label{fig:classify}
\end{figure}

%% file: figs/cls_token.tex
\begin{figure*}[t]
    \centering
    \includegraphics[width=\the\textwidth]{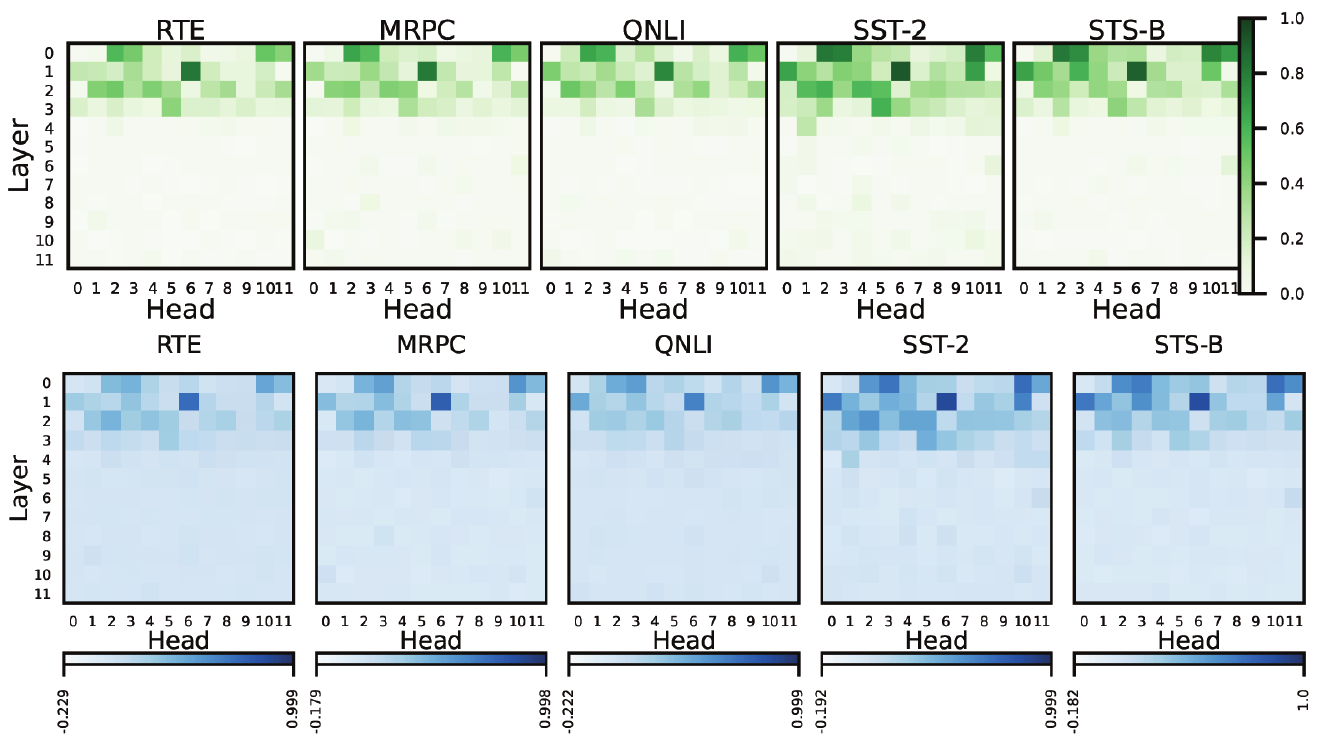}
    \caption{Per-task attention across layers and heads to the \textbf{\clstoken} token when processing other input tokens, averaged over sequence length and dataset items for the selected GLUE task. The darker colors corresponding to larger absolute attention weights. The top (green) figure is computed with the standard attention, and the bottom (blue) figure with the effective attention. Since the effective attention does not have a fixed range as the standard attention (from 0 to 1), we use the minimum and maximum effective attention weight for each task calculated across all weights (not only those associated with the \clstoken token). 
    }
    \label{fig:cls_token}
\end{figure*}

%% file: figs/sep_token.tex
\begin{figure*}[t]
    \centering
    \includegraphics[width=\the\textwidth]{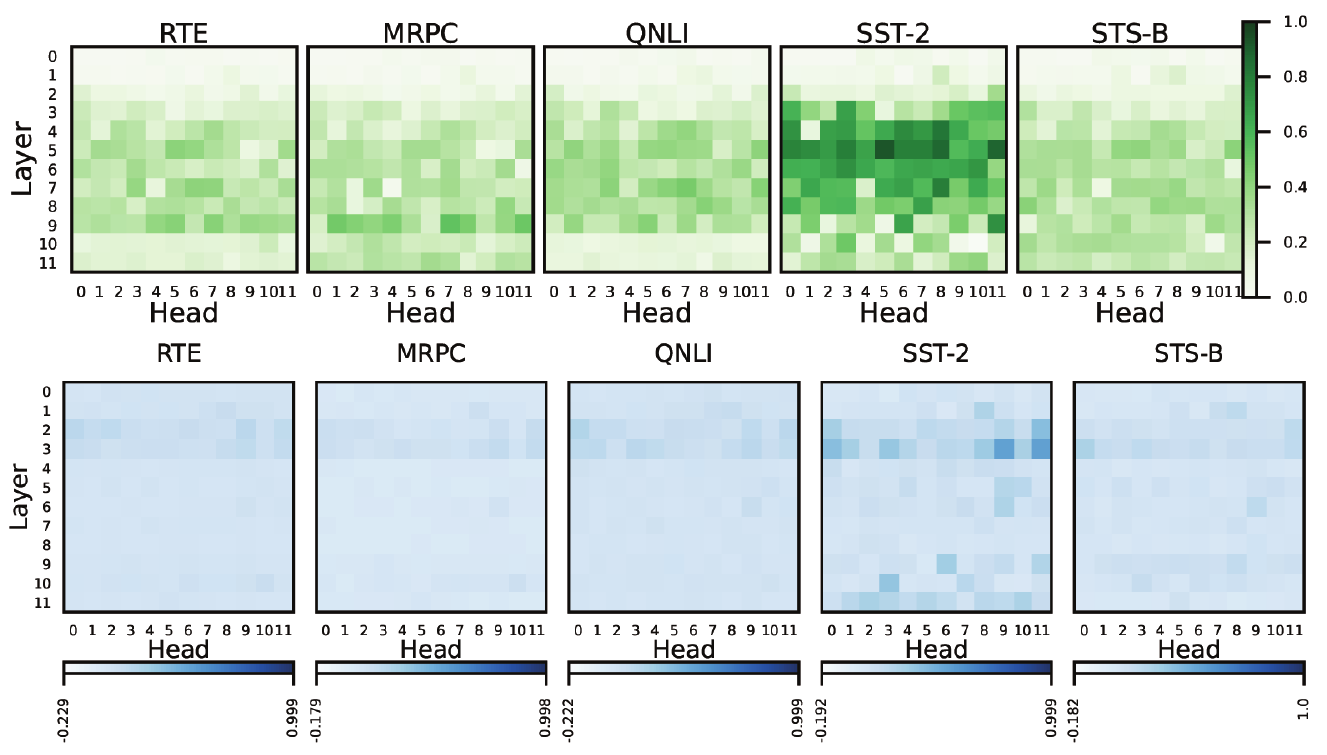} 
    \caption{Per-task attention across layers and heads to the \textbf{\septoken} token when processing other input tokens, averaged over sequence length and dataset items for the selected GLUE task. The darker colors corresponding to larger absolute attention weights. The top (green) figure is computed with the standard attention, and the bottom (blue) figure with the effective attention. Since the effective attention does not have a fixed range as the standard attention (from 0 to 1), we use the minimum and maximum effective attention weight for each task calculated across all weights (not only those associated with the \septoken token). 
    }
    \label{fig:sep_token}
\end{figure*}

%% file: figs/example_figs/example_full.tex
\begin{figure*}[t]
    \centering
    \subfloat[Standard attention]{\includegraphics[width=0.7\textwidth]{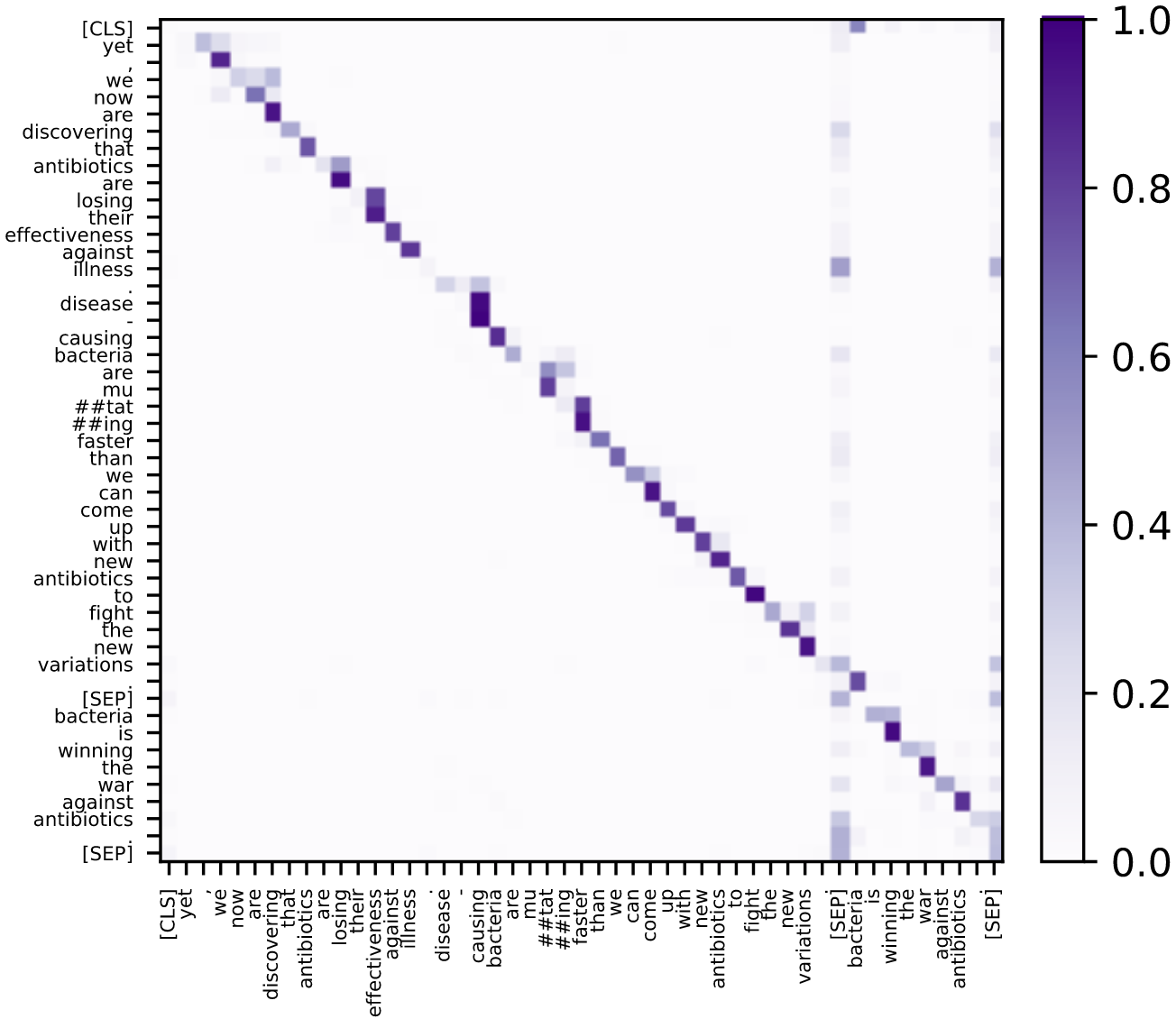} }
    \quad
    \centering
    \subfloat[Effective attention]{\includegraphics[width=0.7\textwidth]{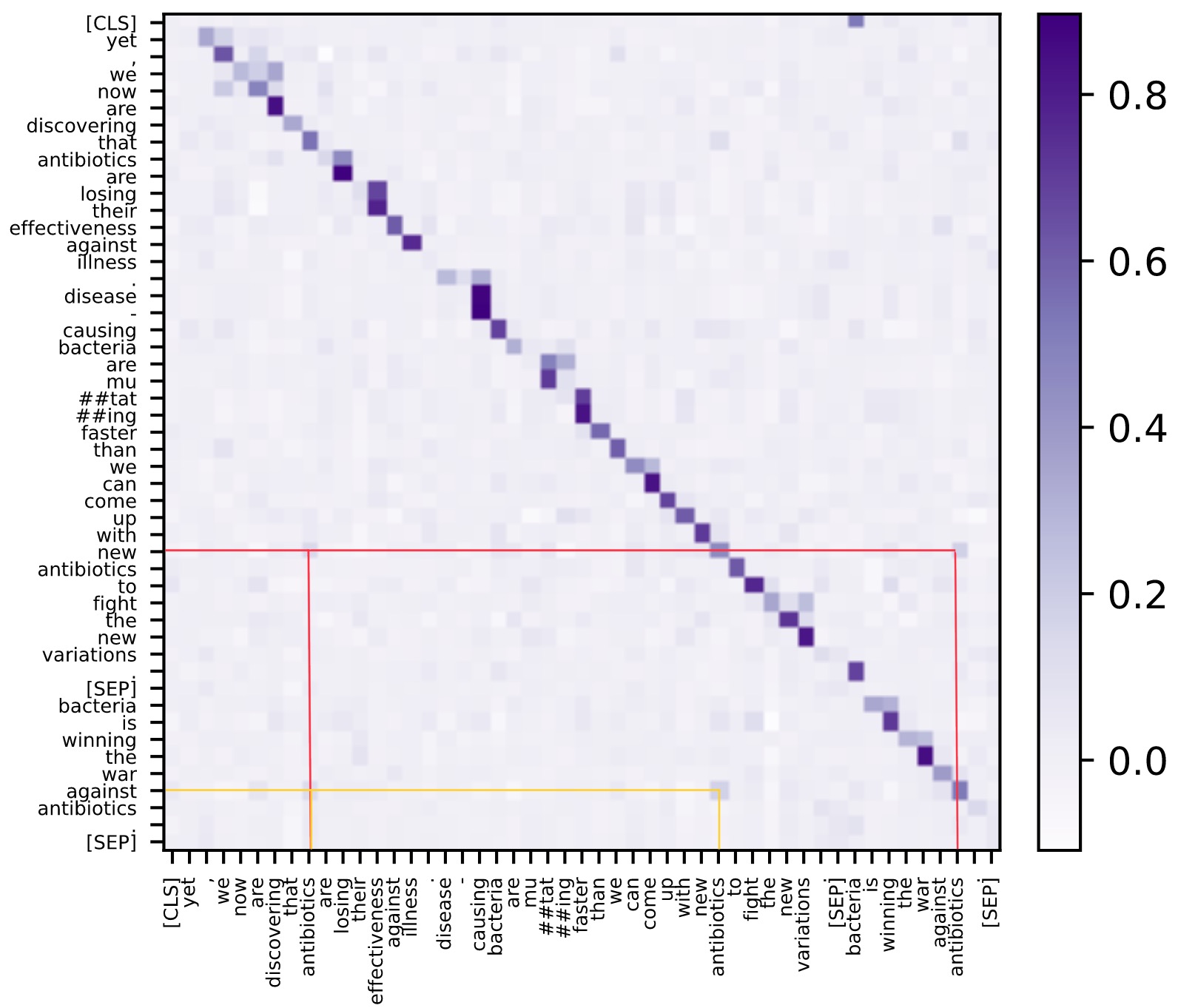} }
    \caption{Complete Figure \ref{fig:example}.}
    \label{fig:example_full}
\end{figure*}